\documentclass[11pt, a4paper]{article}

\usepackage[utf8]{inputenc}
\usepackage[T1]{fontenc}
\usepackage{amsmath}
\usepackage{amsfonts}
\usepackage{graphicx}
\usepackage{hyperref}
\usepackage{url}
\usepackage{times}
\usepackage{titlesec}
\usepackage[backend=biber,style=ieee,sorting=none,url=false]{biblatex}
\usepackage[margin=1in]{geometry}

\titleformat{\section}
  {\normalfont\Large\bfseries\scshape}
  {\thesection}
  {1em}
  {}

\titleformat{\subsection}
  {\normalfont\normalsize\bfseries\scshape}
  {\thesubsection}
  {1em}
  {}

\titleformat{\subsubsection}
  {\normalfont\normalsize\scshape}
  {\thesubsection}
  {1em}
  {}

\title{\textsc{Deep Exploration of Epoch-wise Double Descent in Noisy Data: Signal Separation, Large Activation, and Benign Overfitting}}

\author{
  Tomoki Kubo$^{\dagger}$ \\
  \small Niigata University \\
  \and
  Ryuken Uda$^{\dagger}$ \\
  \small Niigata University \\
  \and
  Yusuke Iida$^{\dagger,\ddagger}$\thanks{To whom correspondence may be addressed. Email: \texttt{iida@ie.niigata-u.ac.jp}} \\
  \small Niigata University
}

\date{}

\begin{document}

\maketitle

\vspace{-2em}
\begin{center}
$^{\dagger}$Graduate School of Science and Technology, Niigata University\\
$^{\ddagger}$Faculty of Engineering, Niigata University
\end{center}
\vspace{1em}

\begin{abstract}
Deep double descent is one of the key phenomena underlying the generalization capability of deep learning models. In this study, epoch-wise double descent, which is delayed generalization following overfitting, was empirically investigated by focusing on the evolution of internal structures. Fully connected neural networks of three different sizes were trained on the CIFAR-10 dataset with 30\% label noise. By decomposing the loss curves into signal contributions from clean and noisy training data, the epoch-wise evolutions of internal signals were analyzed separately. Three main findings were obtained from this analysis. First, the model achieved strong re-generalization on test data even after perfectly fitting noisy training data during the double descent phase, corresponding to a ``benign overfitting'' state. Second, noisy data were learned after clean data, and as learning progressed, their corresponding internal activations became increasingly separated in outer layers; this enabled the model to overfit only noisy data. Third, a single, very large activation emerged in the shallow layer across all models; this phenomenon is referred as ``outliers,'' ``massive activations,'' and ``super activations'' in recent large language models and evolves with re-generalization. The magnitude of large activation correlated with input patterns but not with output patterns. These empirical findings directly link the recent key phenomena of ``deep double descent,'' ``benign overfitting,'' and ``large activation'', and support the proposal of a novel scenario for understanding deep double descent.
\end{abstract}
\section{Introduction}

Artificial intelligence technologies have undergone remarkable development in recent years, introducing substantial transformation to social structures and influencing various academic fields. Although these models form the core of such technologies, the fundamental principles underlying their high generalization capability when trained on real-world data remain poorly understood. Recent numerical experiments have empirically revealed various intriguing phenomena related to this gap. Among them, ``double descent,'' a phenomenon wherein the test error decreases again beyond the overfitting region with increasing number of parameters, is crucial for understanding the generalization capabilities of deep learning models (Belkin et al.~\cite{belkin2019, belkin2020}, Nakkiran et al.~\cite{nakkiran2021a}, Shi et al.~\cite{shi2024}). This phenomenon contradicts the fundamental principle of model construction that excessive parameterization degrades generalization performance; therefore, it has garnered considerable research interest in recent years (d\textquotesingle Ascoli et al.~\cite{ascoli2020}, Chen et al.~\cite{chen2021}, Bodin et al.~\cite{bodin2021}, Nakkiran et al.~\cite{nakkiran2021b}).

Double descent has been reported in linear models. In their early work, Vallet et al.~\cite{vallet1989} observed re-generalization behavior in minimum-norm linear regression using synthetic data. Subsequently, Belkin et al.~\cite{belkin2019} reported similar behavior via numerical experiments on machine learning models, including neural networks, and termed it as ``double descent.'' Nakkiran et al.~\cite{nakkiran2021a} reported that double descent robustly occurs across various tasks and model architectures. They identified model-wise double descent that depends on model parameters and epoch-wise double descent that occurs during model training. After an initial overfitting phase with an increase in test error, continued training leads to a subsequent decrease in test error in the latter phenomenon. They reported that double descent becomes more pronounced with increasing proportion of label noise.

In general, the generalization behavior of machine learning models has been elucidated based on the bias–variance trade-off (Gemam et al.~\cite{geman1992}). Within this framework, it is assumed that once a model's complexity exceeds a certain threshold and enters a low-bias, high-variance overfitting regime, further increases in complexity will not improve its generalization performance. However, recent studies have reported numerous instances across machine learning, wherein such a simplistic understanding is insufficient to explain overparameterized models (Yang et al.~\cite{yang2020} , Adlam et al.~\cite{adlam2020}, Chen et al.~\cite{chen2024}). This has necessitated the development of new approaches for understanding the generalization behavior of machine learning models. As such, linear models have been used in various numerical experiments and theoretical analyses owing to its simplicity. These investigations have provided important insights into double descent, yielding two main scenarios to explain this phenomenon.

The first scenario involves the divergence of the generalization error. Holzmüller~\cite{holzmüller2021} theoretically derived a lower bound for the generalization error of ridgeless linear regression trained on data with noise introduced into the output signal. They showed that the generalization error diverges as the model approaches the interpolation threshold, wherein the number of parameters equals the number of samples; this indicates that model-wise double descent can occur universally. In addition, Mei et al.~\cite{mei2022} performed a theoretical analysis of a two-layer neural network with a fixed random initialization in the first-layer weight and introduced ridge regularization. They rigorously derived the test error curve and showed that under a small ridge regularization coefficient, the test error diverges or increases near the interpolation threshold and causes double descent. Theoretical analyses of error curves for high-dimension ridgeless linear regressions and double descent in logistic regressions have demonstrated divergence in test error at the interpolation threshold (Hastie et al.~\cite{hastie2022}, Bach~\cite{bach2024}, Deng et al.~\cite{deng2021}).

The second explanation concerns the model behavior in response to noise in training data. Ullah et al.~\cite{ullah2024} conducted numerical experiments on double descent caused by noisy data, demonstrating that noise produces an effect equivalent to implicit ridge regularization. Pezeshki et al.~\cite{pezeshki2022} analyzed epoch-wise double descent in linear models and theoretically derived that it occurs due to temporal differences in feature learning, where features with larger variance are learned faster. Similarly, Stephenson et al.~\cite{stephenson2021} demonstrated via theoretical analysis that label noise and differences in feature learning rates can cause epoch-wise double descent in linear models.

Recent studies have further extended numerical experiments and theoretical analyses to deep learning models. For instance, Heckel et al.~\cite{heckel2021} performed numerical experiments on convolutional neural networks (CNNs) and argued that epoch-wise double descent may result from the superposition of multiple bias–variance trade-offs arising from learning different parts of the network at different speeds. In addition, Gamba et al.~\cite{gamba2023} demonstrated that in the overparameterized region following the double descent peak, the curvature of the loss landscape decreases across various models such as CNNs, residual networks (ResNets), and Transformers. They speculated that overparameterization might cause the model to converge to a flatter solution, resulting in a smoother and simpler model function.

Beyond double descent, several interesting phenomena underlying the generalization behavior have been reported such as grokking and benign overfitting.

Grokking is a phenomenon in which prolonged model training suddenly improves the generalization performance from a state of poor generalization despite perfectly fitting the training data. Power et al.~\cite{power2022} reported grokking in a pure mathematical task using Transformers; since then, this phenomenon has been reported in various studies such as in image classification tasks using CNNs and ResNets performed by Humayun et al.~\cite{humayun2024} and pre-training of large language models (LLMs), as reported by Li et al.~\cite{li2025}. In addition, Liu et al.~\cite{liu2023} demonstrated via numerical experiments that grokking can be explained based on differences in learning speed: when the training loss the test loss landscapes are mismatched, large norm values for weight initializations emerge and delay the model convergence to a generalized solution. Recent studies have aimed to elucidate double descent and grokking in a unified framework. For instance, Davies et al.~\cite{davies2022} proposed that these phenomena arise from differences in the feature learning speed, suggesting that epoch-wise double descent and grokking may represent whether heuristic patterns are learned early or late during training. Borkar~\cite{borkar2025} explained these phenomena as two distinct dynamics of stochastic grading descent operating at different timescales. In recent years, frameworks that can handle double descent, grokking, and emergent abilities in a unified manner have also been proposed (Wilson et al.~\cite{wilson2025}, Huang et al.~\cite{huang2024}).

Benign overfitting is a phenomenon in which a model achieves strong generalization performance while perfectly fitting the training data. Neyshabur et al.~\cite{neyshabur2018} empirically observed that the test error continued to decrease even with increasing number of parameters in ResNets. They also derived a generalization bound for two-layer neural networks via theoretical analysis and found that test error decreased with increasing number of hidden units. In addition, Bartlett et al.~\cite{bartlett2020} characterized benign overfitting in linear regression with excessive parameters using the effective rank of the data covariance matrix via theoretical analysis. Tsigler et al.~\cite{tsigler2023} observed benign overfitting when fitting linear basis function models to noisy data. Xu et al.~\cite{xu2023} theoretically demonstrated that a two-layer rectified linear unit (ReLU) network learning noisy exclusive or (XOR) cluster data exhibited benign overfitting and grokking. Other analyses targeting linear models and CNNs have also been conducted, and attempts have been made to enhance the theoretical comprehensions (Cao et al.~\cite{cao2022}, Frei et al.~\cite{frei2023}, Kou et al.~\cite{kou2023}, Jones et al.~\cite{jones2025}).

These studies offer important insights; however, the generalization mechanism underlying deep double descent has not been fully understood. In particular, when training data contain noise, which is significant in real-world tasks, no unified consensus is achieved on the underlying mechanisms. The aforementioned scenarios corresponding to differences in learning speeds and regularization induced by noisy data or model architecture cannot explain the high generalization performance when noisy data are perfectly fitted. Benign overfitting can explain this state, it remains unclear whether a universal mechanism exists for deep learning models to achieve benign overfitting with noisy data. Thus, a new mechanism is required to explain deep double descent when using noisy data.

To address this gap, a novel mechanism underlying deep double descent with noisy data has been investigated herein. Based on the findings from recent studies, we employ traditional, simple model architectures and empirically analyze the epoch-wise evolution of internal structures during training with noisy data; this aspect has not been thoroughly explored in previous studies.
\section{Experimental Setup}

Following the experimental setup reported by Nakkiran et al.~\cite{nakkiran2021a}, the CIFAR-10 dataset~\cite{krizhevsky2009} was used herein; this dataset contains $32 \times 32$ RGB images, including 50,000 training images and 10,000 test images. Label noise was independently introduced for each training sample by randomly changing its label to a class, different from the original one, with a probability of 30\%. Thus, approximately 70\% of training data remained clean, and the remaining 30\% were corrupted by noise.

In addition, three distinct multilayer perceptron (MLP) architectures, which are fully connected neural networks with different number of layers and neuron counts per layer, were considered. These models are denoted as MLP7, MLP5, and MLP3. In particular, the MLP7 architecture comprised six hidden layers and with $2048$, $2048$, $1024$, $1024$, $512$, and $512$ neurons and one output layer with $10$ neurons, respectively. MLP5 comprised four hidden layers with $2048$, $1024$, $512$, and $512$ neurons and one output layer with $10$ neurons. Lastly, MLP3 comprised two hidden layers with $1024$ and $512$ neurons and one output layer with $10$ neurons. The input data have dimensions of $3 \times 32 \times 32$ (channels $\times$ height $\times$ width), corresponding to 3072 features. All models were trained using the Adam Optimizer~\cite{kingma2015} with a learning rate of $10^{-5}$ and default PyTorch hyperparameters, a batch size of 512, ReLU activation functions~\cite{glorot2011}, and a Softmax function~\cite{bridle1990} applied to the output layer before calculating the cross-entropy loss. MLP7 was trained for $10^5$ epochs, whereas MLP5 and MLP3 were trained for $10^6$ epochs due to their slower convergence.
\section{Training Curve Decomposition: Unveiling Distinct Learning Phases of Epoch-wise Double Descent}

\begin{figure}[h]
    \centering
    \includegraphics[width=1\linewidth]{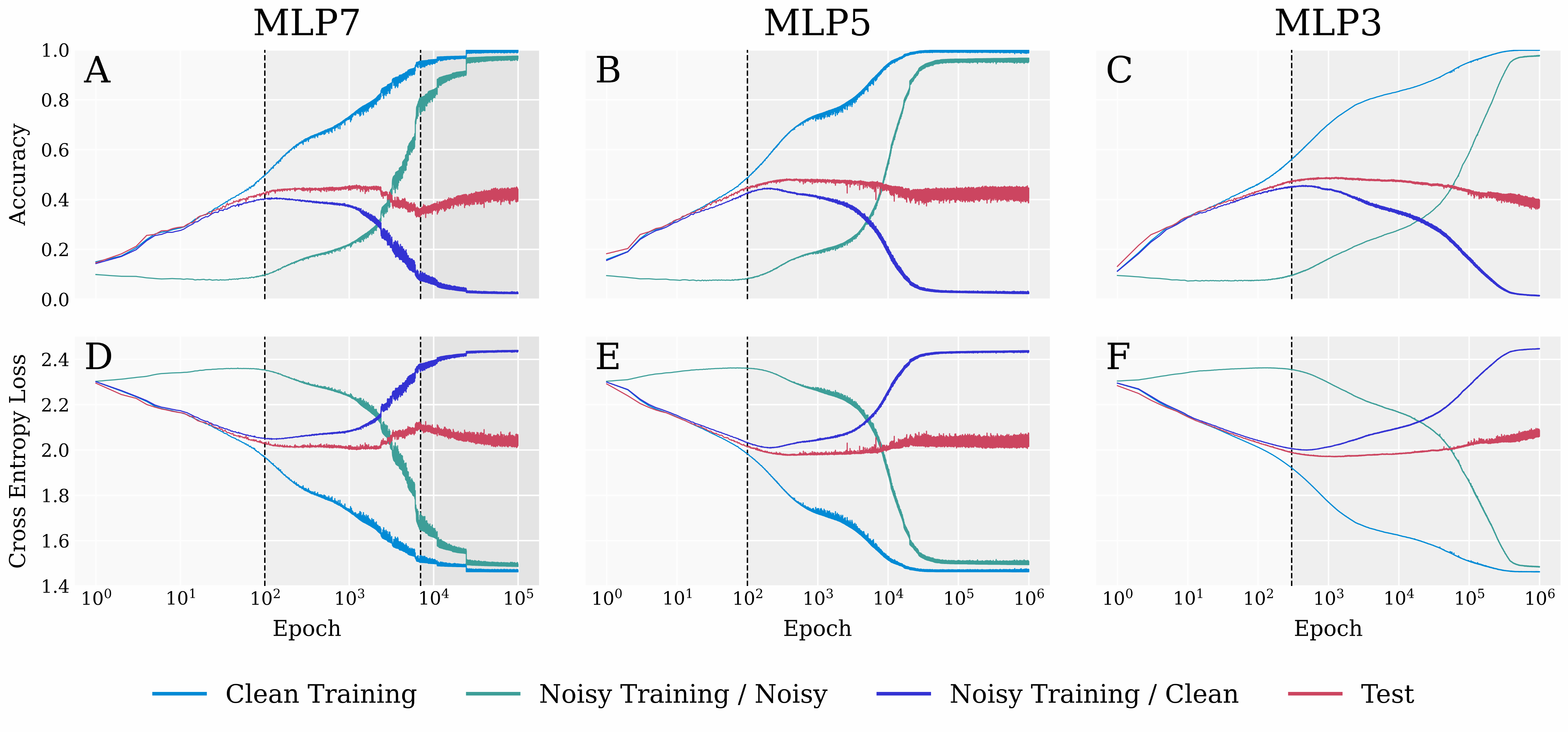}
    \caption{
        Accuracy and loss curves during training. (A)--(C) Accuracy curves and (D)--(F) loss curves. Each column represents a different model: (A) and (D) for MLP7, (B) and (E) for MLP5, and (C) and (F) for MLP3. Training curves are decomposed into two components: ``clean training'' and ``noisy training''. The accuracy and loss values are evaluated on clean training data during ``clean training'' and on noisy training data during ``noisy training.'' ``Noisy training'' is further evaluated using two different label sets: assigned noisy labels (``noisy training / noisy'') and original clean labels (``noisy training / clean''). Vertical dashed lines and shaded background  delineate distinct training phases from their behaviors. Specifically, training phase for MLP7 is divided into three phases: an initial phase where the model fits only clean training data (1--100 epochs), a middle phase where the model fits noisy training data and clean training data, resulting in overfitting (100--7,000 epochs), and a final phase where double descent occurs (7,000--100,000 epochs). In contrast, those for MLP5 and MLP3 are divided into two phases: an initial phase (1--100 epochs for MLP5 and 1--300 epochs for MLP3) and a final phase (100--1,000,000 epochs for MLP5 and 300--1,000,000 epochs for MLP3). 
    }
    \label{fig:loss_acc}
\end{figure}

First, the epoch-wise evolutions of MLP7, MLP5, and MLP3 were analyzed by separately evaluating the loss and accuracy curves on clean and noisy data.

Fig.~\ref{fig:loss_acc} shows the accuracy curves (Fig.~\ref{fig:loss_acc} \textit{A–C}) and loss curves (Fig.~\ref{fig:loss_acc} \textit{D–F}) for the three models: MLP7 (left column; Fig.~\ref{fig:loss_acc} \textit{A} and \textit{D}), MLP5 (middle column; Fig.~\ref{fig:loss_acc} \textit{B} and \textit{E}), and MLP3 (right column; Fig.~\ref{fig:loss_acc} \textit{C} and \textit{F}). 
These curves were decomposed into two main components: ``clean training'' and ``noisy training.'' The clean training component was evaluated on clean training data with their clean labels, whereas the noisy training component was evaluated on noisy training data using two label sets: the assigned noisy labels (``noisy training / noisy'') and the original clean labels (``noisy training / clean'').

An analysis of the loss curve of the ``test'' data showed that MLP7 exhibited double descent per epoch, whereas MLP5 and MLP3 did not. In MLP7, the test loss began decreasing after around 7,000 epochs due to double descent, as indicated by the second vertical dotted line in Fig.~\ref{fig:loss_acc} \textit{A}). MLP5 did not show an increase in test loss, whereas MLP3 experienced an increase in test loss due to overfitting. These behaviors indicated that MLP7 corresponds to the large model regime, MLP5 to the  intermediate regime, and MLP3 to the small model regime, consistent with those reported by Nakkiran et al.~\cite{nakkiran2021a}.

The epoch-wise evolutions of losses on clean and noisy training data were examined. By comparing ``clean training'' and ``noisy training / noisy'' in each panel, their distinct learning phases were identified. The loss curve for MLP7 indicates that the training proceeds in three phases until double descent occurs: (1) an initial phase in which only clean label data are learned up to 100 epochs; (2) an intermediate phase in which data with noisy labels are learned, which increases the test loss up to 7,000 epochs; and (3) a final phase in which noisy labels are well fitted and double descent begins. The loss curves for MLP3 and MLP5 exhibit similar phases (1) and (2) although their corresponding epoch ranges differ. In the first phase, clean label data are only learned up to 100 epochs for MLP5 and 300 epochs for MLP3. This early learning phase in which noisy training data are not learned has also been reported in previous studies (Arpit et al.~\cite{arpit2017}, Liu et al.~\cite{liu2020, liu2022}, Priebe et al.~\cite{priebe2022}). In the subsequent phase, noisy training data are learned, and the models experience the stagnation or an increase of test loss without double descent.

Then, the model state at the end of training was examined. The loss curves for ``clean training'' and ``noisy training / noisy'' (Fig. \ref{fig:loss_acc} \textit{D–F}) show that all models achieved a comparable fit to both clean and noisy training data by the end of training. This result directly indicates that deep double descent observed when using noisy data is caused by the benign overfitting state. However, note that in MLP7, double descent began before the training data were fitted to the upper value of the model even though the model fitted both clean and noisy training data well. This indicates that strong generalization occurs via a mechanism that does not require fitting to the upper limit of the model. In addition, MLP5 and MLP3 fitted noisy and clean training data to the same degree as MLP7, indicating that the perfect fitting of both clean and noisy training data is not sufficient to achieve the benign overfitting state.

This analysis reveals the epoch-wise evolution of how clean and noisy training data are learned and shows that the model exhibits the benign overfitting state even with well fitted noisy training data. However, fitting both clean and noised labels for the same class images can be difficult when signals in the hidden layers pass along the same or at least similar paths to the output layer. Therefore, the separation of the internal signals corresponding to clean and noisy training data was investigated.
\section{Epoch-wise Evolution of Activation Similarity in Hidden Layers: How to Learn Both Clean and Noisy Data Perfectly}

To investigate the separation of internal signals, we analyze the epoch-wise evolution and layer dependence of similarities between the mean activations of hidden layers from clean and noisy training data.

\begin{figure}[t]
    \centering
    \includegraphics[width=0.75\linewidth]{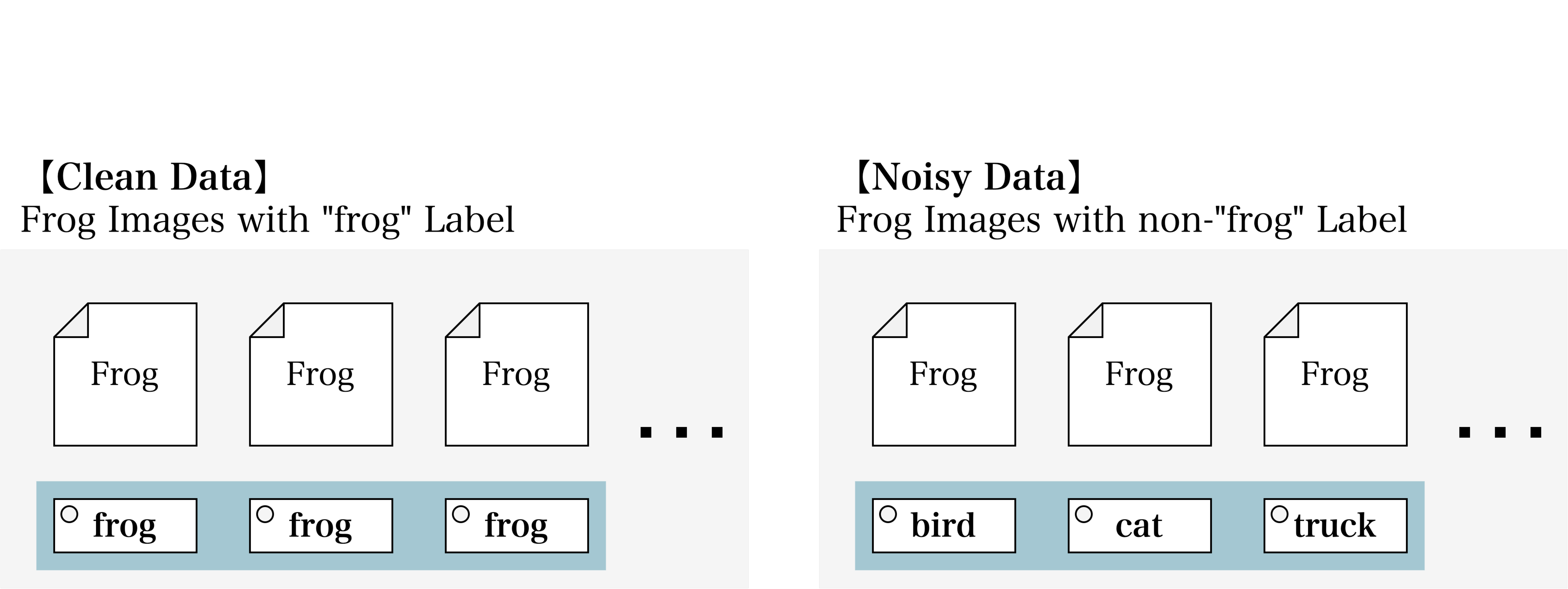}
    \caption{
        Data grouping methods for cosine similarity analysis using the ``frog'' class as an example. ``Clean data'' contain frog images labeled as ``frog,'' whereas ``noisy data'' contain frog images with non ``frog'' labels.
    }
    \label{fig:data_grouping}
\end{figure}

First, the method for grouping data into clean and noisy data employed to compute mean activation magnitudes is described herein. Fig.~\ref{fig:data_grouping} schematizes this data grouping method using the ``frog'' class as an example. ``Clean data'' contain frog images correctly labeled as ``frog,'' whereas noisy data contain frog images with non``frog'' labels. The cosine similarity was employed to quantify the similarity between the mean activations of each hidden layer from clean and noisy training for each class. The cosine similarity of class $c$ at layer $l$, $CS^{(l)}_{c}$, is defined as follows:

\begin{align}
    CS^{(l)}_{c} &= \frac{\boldsymbol{S}^{(l)}_{\text{clean},c} \cdot \boldsymbol{S}^{(l)}_{\text{noisy},c}}{\|\boldsymbol{S}^{(l)}_{\text{clean},c}\| \|\boldsymbol{S}^{(l)}_{\text{noisy},c}\|} \, , \label{eq:cosine_similarity} \\
    \boldsymbol{S}^{(l)}_{\text{clean},c} &= \frac{1}{n_\text{clean,c}} \sum^{n_\text{clean,c}}_{m=1} \text{model}(d_{\text{clean},c,m})^{(l)} \, , \label{eq:clean_signal} \\
    \boldsymbol{S}^{(l)}_{\text{noisy},c} &= \frac{1}{n_\text{noisy,c}} \sum^{n_\text{noisy,c}}_{m=1}
    \text{model}(d_{\text{noisy},c,m})^{(l)} \, , \label{eq:noisy_signal}
\end{align}

\begin{figure}[t]
    \centering
    \includegraphics[width=1\linewidth]{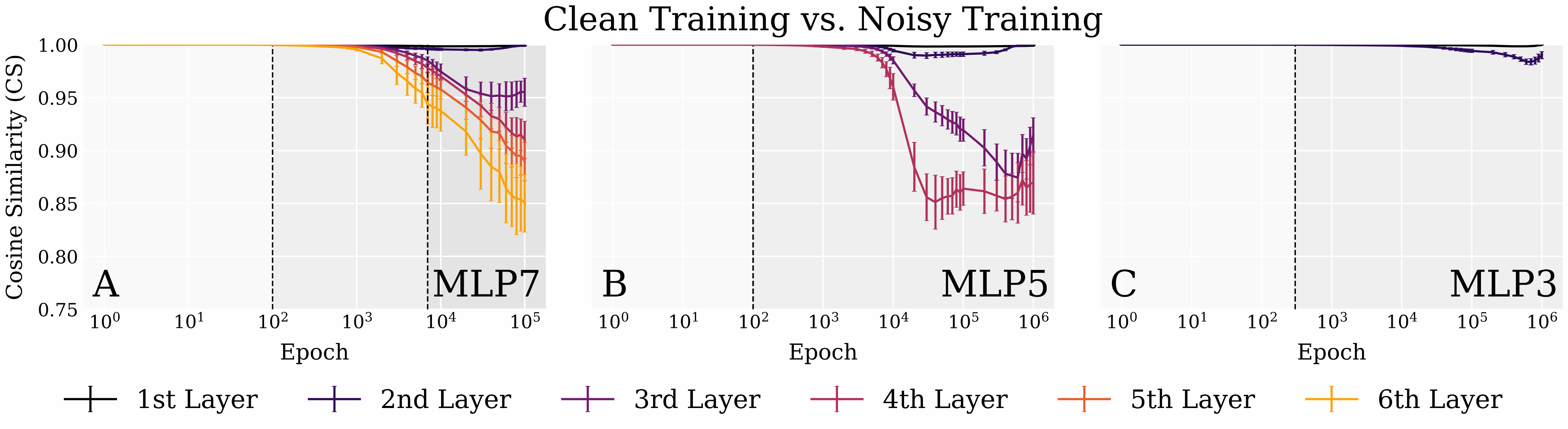}
    \caption{
        Epoch-wise evolution of cosine similarity between the mean activations of hidden layers from clean and noisy training data for MLP7, MLP5, and MLP3 from left to right. Vertical dashed lines and background colors indicate distinct training phases, consistent with those in Fig.~\ref{fig:loss_acc}. The hidden-layer activations become increasingly separable in outer layers as the training progresses. MLP3 shows a weaker degree of separation, whereas MLP7 and MLP5 show comparable degrees of separations in the outer layers.
    }
    \label{fig:clean_noised}
\end{figure}

where $n_{\text{clean},c}$ and $n_{\text{noisy},c}$ denote the amount of clean training data and noisy training data belonging to class $c$, $\text{model}(d_{\text{clean},c,m})^{(l)}$ and $\text{model}(d_{\text{noisy},c,m})^{(l)}$ denote the signals at layer $l$ when the $m$-th clean training data and noisy training data of class $c$ are input into the models, and $\boldsymbol{S}^{(l)}_{\text{clean},c}$ and $\boldsymbol{S}^{(l)}_{\text{noisy},c}$ denote the mean activation signals. $CS^{(l)}_{c}$ denotes the cosine similarity of class $c$ at layer $l$. The average cosine similarities across all 10 classes for each layer were computed as follows:

\begin{align}
    CS^{(l)} = \frac{1}{10} \sum^{10}_{c=1}CS^{(l)}_c \, . \label{eq:cosine_similarity_avg}
\end{align}

This process was repeated for each epoch sampled on a logarithmic scale over training steps, i.e., $n \times 10^m$ ($n, m \in \mathbb{N}$).

\begin{figure}[htbp]
  \centering
  \begin{minipage}{1\columnwidth}
    \centering
    \includegraphics[width=1\linewidth]{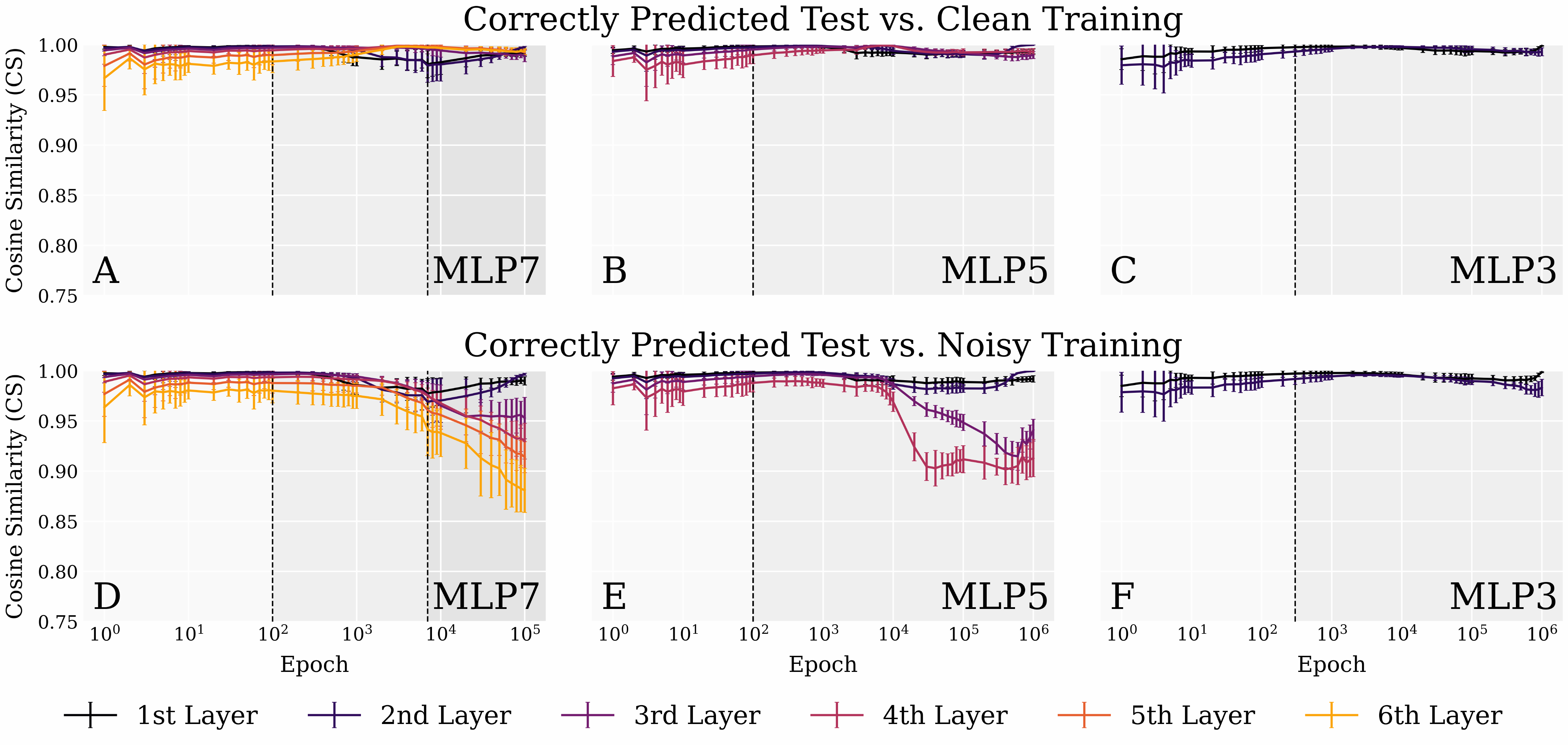}
    \caption{
        Epoch-wise evolution of cosine similarity between the mean activations of hidden layers for correctly predicted test data and clean training data (top row) and those for correctly predicted test data and noisy training data (bottom row). Each column represents a different model: (A) and (D) are for MLP7, (B) and (E) for MLP5, and (C) and (F) for MLP3. The correctly predicted test data signals are closely aligned with those of clean training data signals for all models, whereas these are separated from noisy training data signals in MLP7 and MLP5.
    }
    \label{fig:correct_test}
  \end{minipage}
  
  \vspace{10pt}
  
  \begin{minipage}{1\columnwidth}
    \centering
    \includegraphics[width=1\linewidth]{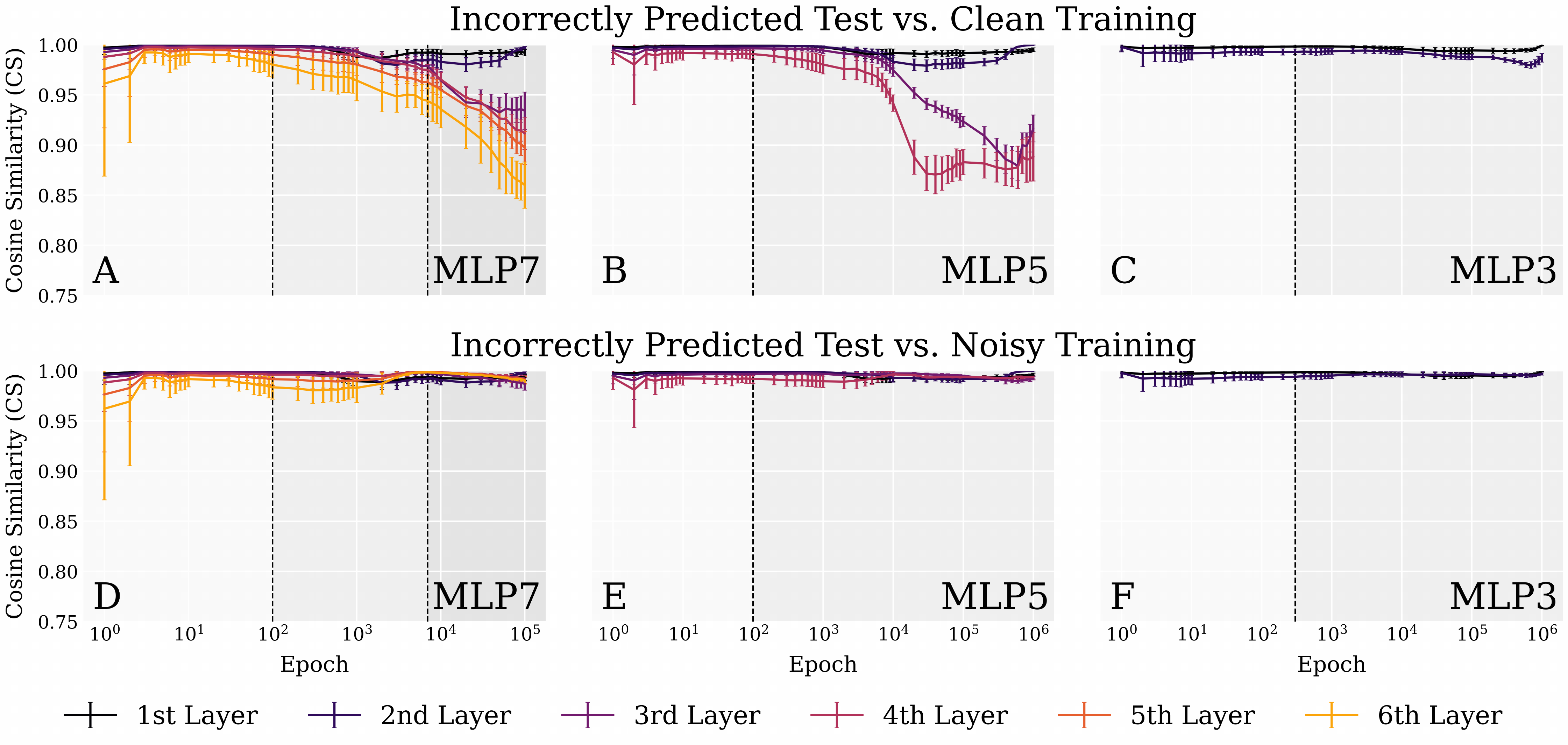}
    \caption{
        Same as Fig.~\ref{fig:correct_test} but for incorrectly predicted test data.
    }
    \label{fig:incorrect_test}
  \end{minipage}

\end{figure}

Fig.~\ref{fig:clean_noised} shows the epoch-wise evolution of cosine similarity for MLP7, MLP5, and MLP3. Error bars represent the standard deviation between classes. The output layer is excluded. The activations for clean and noisy data became increasingly separable as the training progressed, as evidenced by a decrease in cosine similarity. All three models exhibited this behavior. In addition, MLP7 and MLP5 showed comparable degree of separability, whereas MLP3 showed a lower degree of separation. The separation was particularly pronounced in deeper layers, as indicated by light-colored lines in the figures that corresponded to the layers closer to the output. The epochs at which separation begins for each model also exhibit distinct characteristics. In the curve for MLP7, the separation begins considerably earlier than the second vertical dotted line marking the onset of double descent in the loss curve. A comparison of Fig.~\ref{fig:clean_noised} with the loss curves in Fig.~\ref{fig:loss_acc} shows the onset of separation coincides with the epochs at which the ``noisy training / noisy'' loss becomes steep, i.e., $10^3$ epochs for both MLP7 and MLP5 and $3 \times 10^4$ epochs for MLP3. These results indicate that signal separation occurs when the model effectively learns noisy training data well and that deeper models can be more effective at separating clean and noisy data signals.

This analysis was also performed on test data to investigate whether signal separation occurs and determine it effect on the generalization performance of the model. The cosine similarities between the mean hidden-layer activations of test data and clean training data, as well as between test data and noisy training data, were investigated. This analysis was performed separately for correctly and incorrectly predicted test samples because their signal propagation paths are expected to differ. Note that the sets of correctly and incorrectly predicted test data vary at each epoch.

Fig.~\ref{fig:correct_test} shows the epoch-wise evolution of cosine similarity between correctly predicted test data and clean training data (Fig.~\ref{fig:correct_test} \textit{A–C}) / noisy training data (Fig.~\ref{fig:correct_test} \textit{D–F}) across the three models: MLP7 (left column; Fig.~\ref{fig:correct_test} \textit{A} and \textit{D}), MLP5 (middle column; Fig.~\ref{fig:correct_test} \textit{B} and \textit{E}), and MLP3 (right column; Fig.~\ref{fig:correct_test} \textit{C} and \textit{F}). Fig.~\ref{fig:incorrect_test} shows the same plot but for the incorrectly predicted test data. The vertical dashed lines in each panel indicate transitions between learning phases and consistent with those shown in Fig.~\ref{fig:loss_acc}.

Fig.~\ref{fig:correct_test} shows that for MLP7 and MLP5, the internal signals of the correctly predicted test data exhibit large cosine similarities, i.e., $\sim 1$,  with those of clean training data throughout training and remain separated from those of noisy training data. A comparison of these results with Figs.~\ref{fig:clean_noised} \textit{A} and \textit{B} shows that the epoch-wise evolution of separation from the noisy training data closely aligns with that observed between clean and noisy training data. The separation increases as the training proceeds and becomes more pronounced in deeper layers. However, the degree of separation between correctly predicted test data and noisy training data is smaller than that observed between clean and noisy training data. In contrast, Fig.~\ref{fig:incorrect_test} reveals a different behavior for incorrectly predicted data. The internal signals exhibit high similarity with that of noisy training data but low similarity with that of clean training data. These results indicate that the correctly and incorrectly predicted test data are processed via distinct pathways within the models, corresponding to the separated signals of clean and noisy training data respectively, in MLP7 and MLP5. In contrast, the mean activations of training and test data exhibit high similarity throughout training, as shown in Fig.~\ref{fig:correct_test} \textit{C}, \textit{F} and Fig.~\ref{fig:incorrect_test} \textit{C}, \textit{F}.

Herein, the degree of signal separability was quantified based on the mean activations of hidden layers of different datasets. Results revealed that the internal signals were clearly separated in MLP7 and MLP5 but not in MLP3. Subsequently, the mean activation signals were analyzed and the corresponding results are discussed in the next section.
\section{Investigation of Activations in the Double Descent Phase: Emergence of Large Activation and its Characteristics}

When training near the onset of double descent, a fixed single neuron exhibited considerably higher activation magnitudes compared with those of other neurons. Similar phenomena have been reported in recent LLMs or Transformer-based models, wherein large activation occurred in the double descent phase with our setup. 

\begin{figure}[t]
    \centering
    \includegraphics[width=1\linewidth]{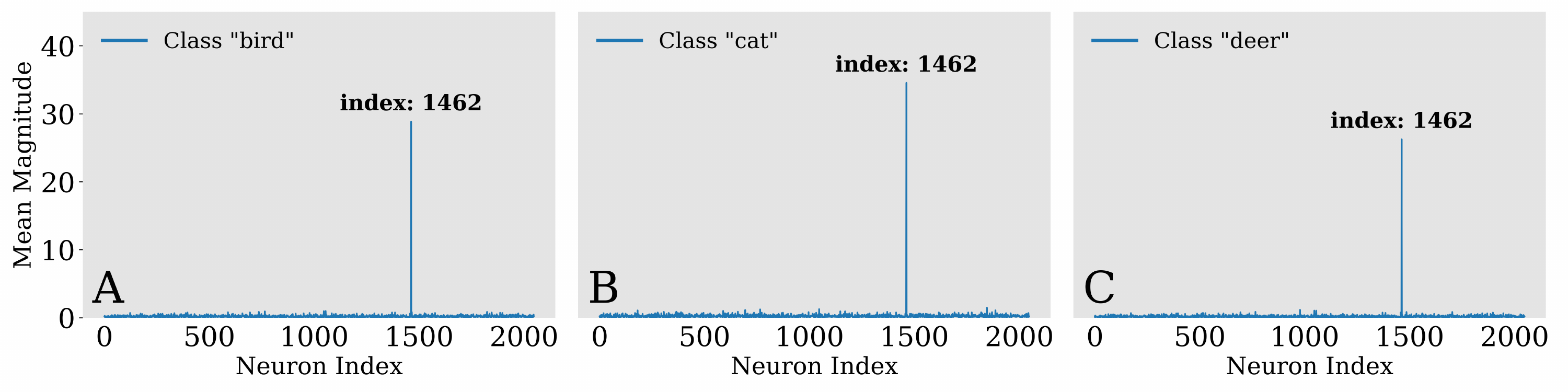}
    \caption{
        Mean activation magnitudes of each neuron in the second hidden layer of MLP7 at 100,000 epochs for a representative class. Activations are averaged over the entire training dataset for inputs belonging to the ``bird,'' ``cat,'' and ``deer'' classes from left to right. The neuron with index 1462 exhibits a remarkably large activation magnitude compared with other neurons, regardless of the input class.
        }
    \label{fig:mlp7_mean_activaions}
\end{figure}

Fig.~\ref{fig:mlp7_mean_activaions} shows the mean activation magnitudes of each neuron in the second layer of MLP7 at 100,000 epochs for representative classes; these magnitudes are averaged over the entire training dataset for each class. From left to right, the panels show examples for the ``bird,'' ``cat,'' and ``deer'' classes. Note that the neuron with index 1462 exhibits a remarkably large activation magnitude compared with other neurons, regardless of the input class. This behavior is not limited to these three classes shown but is consistently observed across all 10 classes. These large activation magnitudes have been referred to as \textit{large activation} herein.

\begin{figure}[t]
    \centering
    \includegraphics[width=1\linewidth]{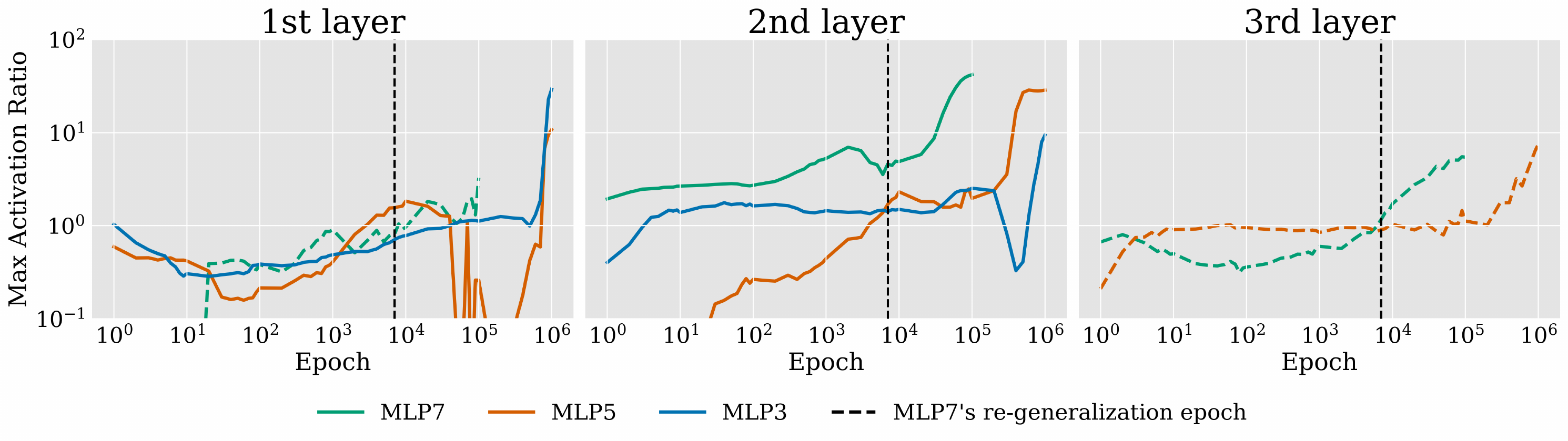}
    \caption{
        Epoch-wise evolution of the ratio of the maximum activation magnitude to the root mean square (RMS) of all remaining activations in the same layer for MLP7, MLP5, MLP3. From left to right, panels correspond to the results for the first, second, and third hidden layers of each model. The ratios are specifically calculated by tracking the neuron that exhibits the largest activation magnitude at the final epoch. If the maximum ratio does not exceed 10, the evolution is shown as a dashed line. The vertical dashed lines indicate the epoch of double descent onset in MLP7 (7,000 epochs). Large activation emerges and become more pronounced in the later stages of training and are observed only in shallow layers, specifically up to the third layer, for all models.
    }
    \label{fig:max_activation_evolution}
\end{figure}

As previously mentioned, similar phenomena such as \textit{outliers}, \textit{massive activation}, and \textit{super activation}(Bondarenko et al.~\cite{bondarenko2021}, Sun et al.~\cite{sun2024}, Yu et al.~\cite{yu2025}) have been observed in Transformer-based models and LLMs. Although multiple neurons exhibit large activation in these models, they occur at fixed feature dimensions independent of the input token or prompt. Yu et al.~\cite{yu2025} further demonstrated that pruning just a single super weight that induces super activation resulted in a significant score drop and an increase in stopwords in the generated text; this performance drop is severe than when pruning thousands of other normal weights. These findings indicate that these highly activated neurons play a critical role in deep learning models and within the context of double descent. The proposed model architecture is simpler than those reported in previous studies, enabling the straightforward analysis of large activation. To investigate the emergence and development of such large activation during training, their magnitude relative to the overall activations was analyzed.

Fig.~\ref{fig:max_activation_evolution} shows the epoch-wise evolution of the maximum activation magnitude in MLP7, MLP5, and MLP3. The vertical axis represents the ratio of the maximum activation magnitude to the RMS of all other activations in the same layer. This ratio is calculated by tracking the neuron that exhibits the largest activation magnitudes at the final epoch. The panels show the results for the first, second, and third layers of each model, from left to right. Note that the third hidden layer is not shown for MLP3 as it is the output layer.
The ratio $r_t$ at epoch $t$ is defined for each layer as follows:

\begin{align}
    r_t &= \frac{a_{t,i^*}}{\sqrt{\frac{1}{m-1}\sum^m_{i=1, i \ne {i^*}} a^2_{t,i}}} \, .
\end{align}

where $a_{t,i}$ denotes the activation magnitude of the $i$-th neuron in this layer at epoch $t$, $i^*$ is the index of the maximum activation neuron at the final epoch, and $m$ is the number of neurons in the layer.

``Large activation'' was observed across all models, which are neurons whose activation ratio exceeds 10; these are indicated by solid lines in Fig.~\ref{fig:max_activation_evolution}. These activations developed and became more pronounced in the later stages of training. In MLP7, indicated by the green solid line in Fig.~\ref{fig:max_activation_evolution}, a ``large activation'' occurred in the second hidden layer. In addition, the epoch at which this large activation develops coincides with the vertical dotted line that marks the onset of double descent in the loss curve shown in Fig.~\ref{fig:loss_acc}. Interestingly, increases in maximum activation were also observed in the first and third layers of MLP7 although they did not reach the threshold. The corresponding results for MLP5 and MLP3 are indicated by orange and blue solid lines, respectively. In both these models, ``large activation'' is observed in the first and second layers just before $10^6$ epochs, which is later than in MLP7. Note that ``large activation'' was found only in the shallow layers, specifically up to the third layer, and not in deeper layers within $10^6$ epochs.

\begin{figure}[t]
    \centering
    \includegraphics[width=1\linewidth]{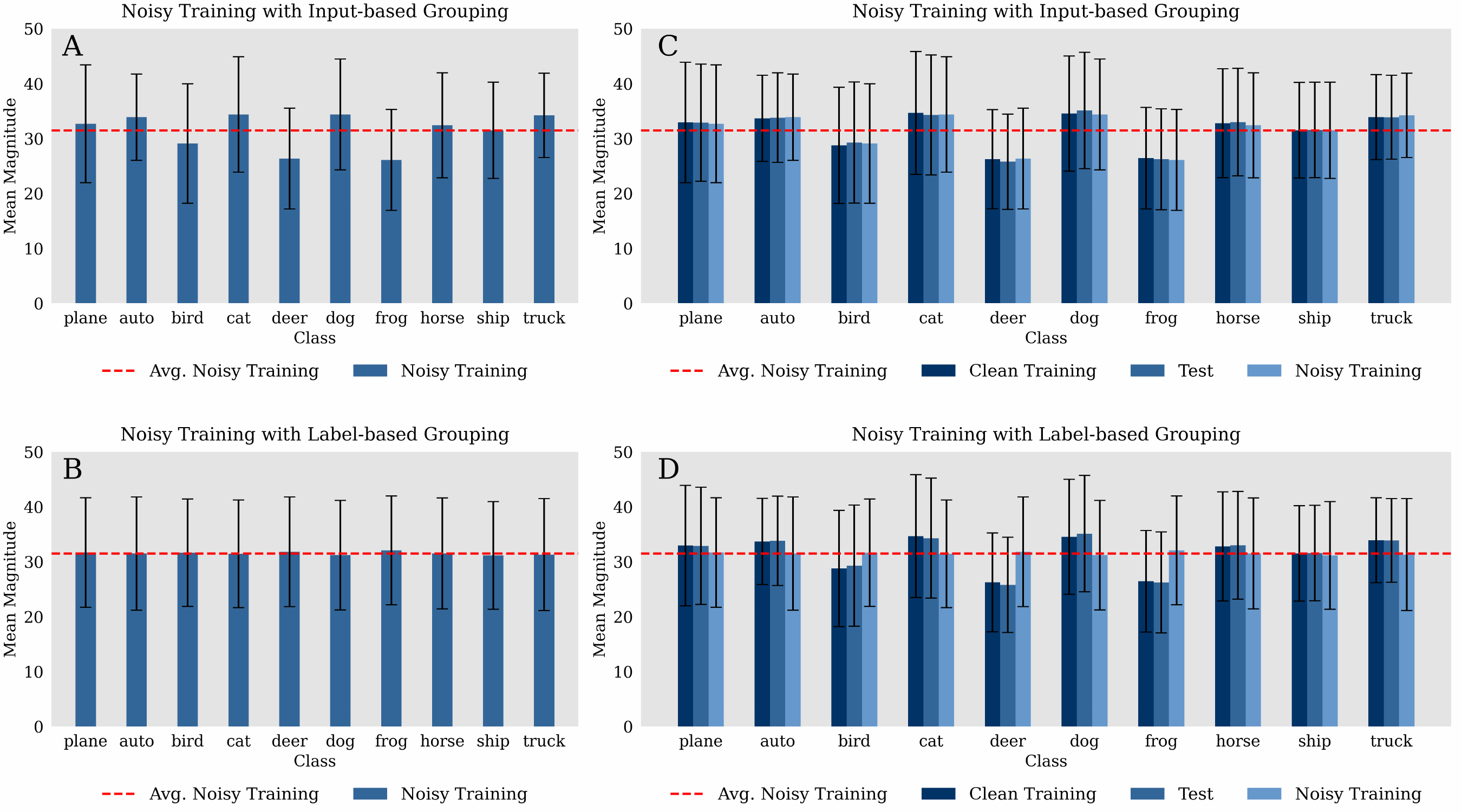}
    \caption{
        Correlation analysis of large activation magnitudes with respect to class labels via input-based or label-based grouping. Each panel shows the mean magnitude of maximum activation in the second hidden layer of MLP7 at the final epoch, i.e., the neuron with index 1462. Error bars indicate the standard deviation across each class. (A) Noisy training data with input-based grouping; (B) noisy training data with label-based grouping; (C) clean training data, test data, and noisy training data with input-based grouping; and (D) clean training data, test data, and noisy training data with label-based grouping. The red dashed lines indicate the mean magnitude of the maximum activation for all noisy training data to facilitate comparison across all figure panels.
    }
    \label{fig:max_activation_per_class}
\end{figure}

To further elucidate the role of ``large activation,'' their correlations with input patterns and output class labels were analyzed. Fig.~\ref{fig:max_activation_per_class} shows the mean magnitude of ``large activation'' for each class in the second hidden layer of MLP7 at the final epoch, i.e., the neuron at index 1462 shown in Fig.~\ref{fig:max_activation_evolution}. The error bars represent the standard deviation across samples. Fig.~\ref{fig:max_activation_per_class} \textit{A} and \textit{B} show the results obtained using input-based and label-based grouping, respectively. Fig.~\ref{fig:max_activation_per_class} \textit{C} and \textit{D} show the results obtained by including clean training data and test data addition to the results in A and B, respectively, using the same grouping method. The horizontal red dotted line denotes the mean magnitude of ``large activation'' for noisy training data.

A comparison of Fig.~\ref{fig:max_activation_per_class} \textit{A} and \textit{B} shows that the mean magnitude of ``large activation'' exhibits dependence on the input pattern although the variation across classes is relatively large, whereas it remains nearly identical across output class labels. These results indicate that the development of ``large activation'' is primarily driven by input pattern information rather than output labels. Then, the ``large activation'' magnitudes across clean and noisy training data and test data are compared; the corresponding results are shown in Fig.~\ref{fig:max_activation_per_class} \textit{C} and \textit{D}. As clean training data and test data have true labels, input-based and label-based groupings are equivalent for these datasets. Moreover, data with the same class labels exhibit similar mean ``large activation'' magnitudes in input-based grouping but not in label-based grouping. These results support the idea that the ``large activation'' plays a role in facilitating strong generalization performance in the double descent state.
\section{Discussion}

\subsection{Summary of Results}

Herein, deep double descent was empirically investigated using traditional, simple model architectures, focusing on two aspects: the presence of data noise and epoch-wise evolution of the internal structure. This analyses yielded three primary results.

\subsubsection*{1. Learning Phase Transition}

In all models, clean training data were learned during the earliest phase, followed by the learning of noisy training data. ``Double descent'' occurred only in MLP7 after it achieved good fitting to both clean and noisy training data. Note that the loss on noisy training data slightly decreased after the onset of double descent.

\subsubsection*{2. Internal Signal Separation}

In MLP7 and MLP5, the cosine similarity between the hidden-layer activations for clean and noisy training data decrease during training and the degree of separation is larger in deeper layers, corresponding to accelerated learning of noisy training data.

\subsubsection*{3. Development of Large Activation}

``Large activation'', also referred to as ``outliers,'' ``massive activations,'' or ``super activations'' in existing LLM-based studies, were observed in all models after learning noisy data. A fixed neuron at the shallower layer exhibited considerably larger activation magnitudes across all input classes. In MLP7, large activation emerged from the epoch at which re-generalization via ``double descent'' appeared in the loss curve. The mean magnitudes of these activations were correlated with the correct class labels in clean training data, noisy training data, and test data.

\subsection{Why Does Internal Signal Separation Occur?}

In MLP5 and MLP7, clear separation was observed between the internal signals of clean and noisy training data. We presume that such separation occurs because it is necessary for the model to produce different outputs from similar input patterns belonging to the same class; these patterns are initially mapped closely during the early learning phase. This interpretation is supported by the observation that the loss associated with noisy training data decreases sharply around the epoch at which internal signal separation occurs. This interpretation is further supported by the analysis results for test data. The representations of ``correctly predicted test and clean training data'' as well as ``incorrectly predicted test and noisy train'' exhibit high similarity. In contrast, other combinations exhibit lower similarity. This indicates the formation of distinct paths for the former combination and the latter combination. In addition, MLP3, where overfitting progresses after the model begins to learn noisy training data, shows considerably weaker degree of separation. This indicates that sufficient separation is necessary for achieving deep double descent with noisy training data.

Gu et al.~\cite{gu2024} performed a similar analysis of model-wise double descent in an image classification task. They found that noisy and clean training samples were positioned close together in the feature space of the final layer, contradictory to the results of the present study. However, their fully connected neural network was considerably smaller than the three models discussed herein, suggesting that signal separation has not progressed sufficiently to enable effective learning of noisy data. This interpretation is further verified in the Supporting Information (\textit{SI Appendix}, Figs. S1 and S2). Gu et al. also reported greater separations in larger CNNs and ResNets, consistent with the findings reported herein.

\subsection{Emergence of Large Activation}

``Large activation,'' in which the same neuron exhibit considerably larger activation magnitudes for certain classes compared to others, were observed across all models. The same phenomena, although with considerably larger activation magnitudes, was reported in previous studies as ``outliers,'' ``massive activations,'' or ``super activation'' (Bondarenko et al.~\cite{bondarenko2021}, Sun et al.~\cite{sun2024}, Yu et al.\cite{yu2025}) in larger and more complex model architectures such as Transformer-based models and LLMs. Surprisingly, such phenomena were confirmed even in our simpler, fundamental deep neural networks architecture. Note that as our model architecture and task are considerably simple than those reported in previous studies, novel properties relevant to the generalization mechanism could be elucidated using a straightforward analytical method. Moreover, the emergence of large activation corresponded to the epoch at which ``double descent'' appeared in the loss curve; this directly confirmed the relevance of large activation to re-generalization. The mean magnitude of ``large activation'' also corresponded to the input class or true label but not to the noisy training labels. This indicated that ``large activation'' arise from the effective compression of input patterns.

\subsection{Attainment of Re-generalized State}

What type of state does ``double descent'' correspond to? Loss curve decomposition clearly revealed that when ``double descent'' occurs, the model achieves a re-generalized state despite good fitting to noisy training data, i.e., a ``benign overfitting'' state. 

Bartlett et al.~\cite{bartlett2020} reported ``benign overfitting'' in linear basis function models when fitting noisy trigonometric function data using considerably higher number of parameters. They reported that benign overfitting emerged when the $n$-th cosine basis function was scaled by $\frac{1}{n}$, emphasizing low-frequency components, while the high-frequency basis function exhibit behavior that complements the noise. 

The results reported herein, particularly the separation of internal activation signals and the emergence of ``large activation,'' indicate a novel and plausible scenario in which a similar mechanism occurs in ``deep double descent'' with noisy data. The separation between the internal signals of clean and noisy training data indicates that a small subnetwork learns the important overall patterns in input training data, whereas the remaining parts of the model overfit the separated noisy data. Thus, the model achieves ``benign overfitting.'' Previous studies have shown that neural networks can learn even purely random labels owing to their high expressive power (Zhang et al.~\cite{zhang2017}), supporting our interpretation. However, a large subnetwork is required to learn noisy training data. To achieve this, a ``large activation'' emerges in the shallow layers to compress input patterns into the smallest possible subnetwork. The internal signal separation mainly occurs after this layer and noisy training data are overfitted in the outer layers, resulting in the ``benign overfitting'' state.

\subsection{Suggestion of a Novel Scenario for ``Double Descent''}

\begin{figure}[t]
    \centering
    \includegraphics[width=0.8\linewidth]{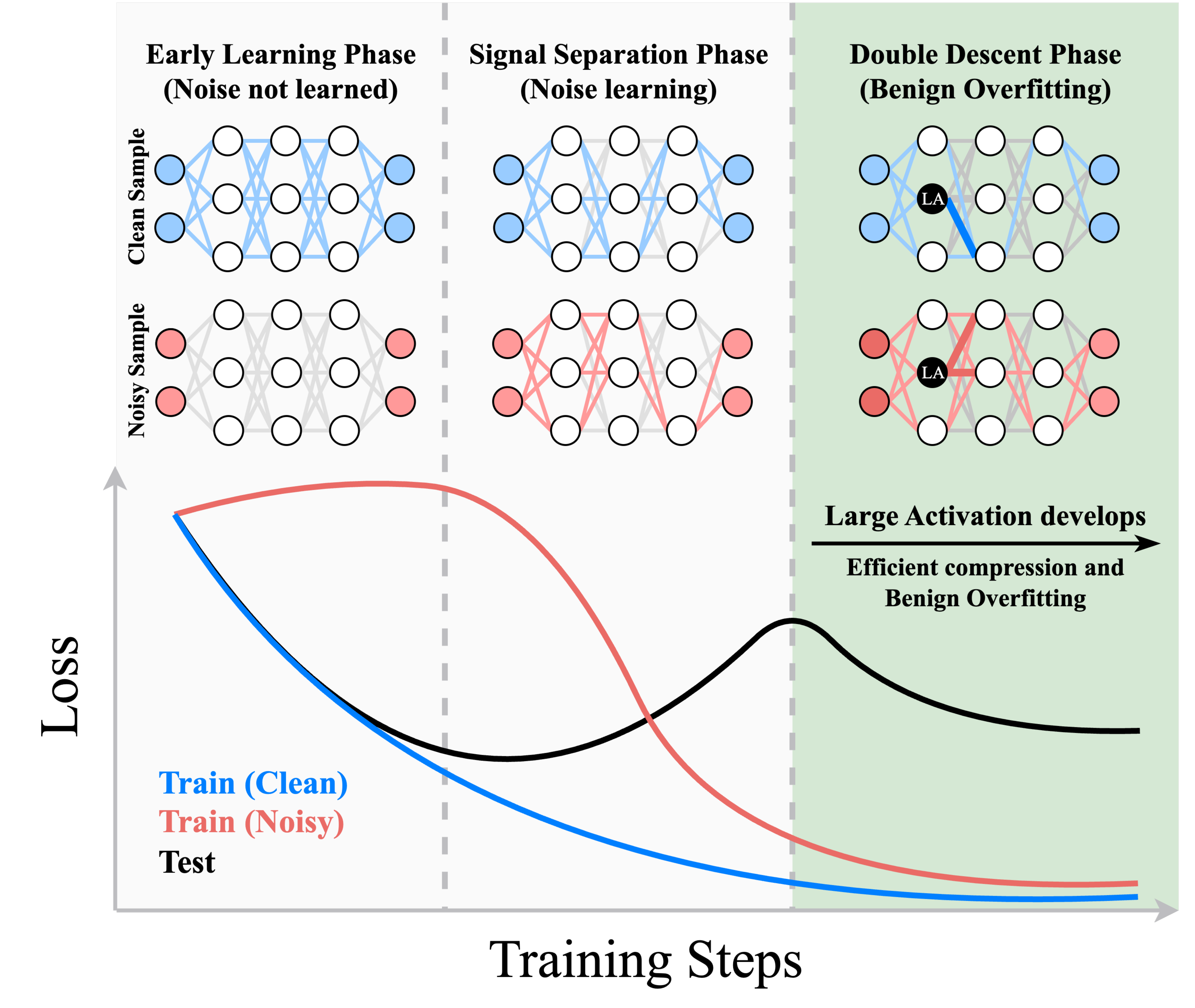}
    \caption{
        The proposed scenario of internal development from the early learning phase until after the onset of double descent. Loss curves for clean training data, noisy training data, and test data are shown at the bottom. Internal states for each phase are also shown at the top, corresponding to when the model is fed with clean or noisy training samples. Early Learning Phase: The model fits only clean training data, whereas noisy training data are not learned. No internal paths are formed to separate noisy training data in this phase. Signal Separation Phase: The model begins to learn noisy training data, increasing the test error. Internal signals for clean and noisy training data begin to separate, forming distinct paths. Double Descent Phase: Large activation emerges and grows, acting as an efficient compression of input patterns. This enables the model to learn the essential characteristics of input data with a small, sparse network. Thus, the remaining sufficiently large network can fully fit the noisy training data, and the model reaches the ``benign overfitting'' state. This triggers a second decrease in the test loss curve, indicating ``double descent.''
    }
    \label{fig:overview}
\end{figure}

Fig.~\ref{fig:overview} shows the proposed scenario and epoch-wise internal evolution of the model from the early learning phase to the ``double descent'' phase in the experimental setup. In the early learning phase, the model only learns clean training data and does not learn noisy training data. After training, the model begins to learn noisy training data; however, learning all noisy data is difficult for the model. Thus, the internal signals corresponding to clean and noisy training data are separated, thereby enabling the model to effectively learn noisy data. As learning progresses further, individual noise data must be fit perfectly to reduce the training loss; this process requires a large subnetwork. To achieve this state, a large activation emerges in the shallow layers; this activation compresses the input pattern into a small, sparse subnetwork. After this suppression, internal signal separation occurs in subsequent layers, wherein noisy training data are overfitted. Consequently, the entire model reaches the ``benign overfitting.'' The proposed scenario is straightforward and may naturally be inherent in models with a large expressive capacity and multiple internal paths, thereby enabling noise to be learned separately from correct data.

\section{Concluding Sentence}

Herein, epoch-wise deep double descent in traditional, simple model architectures was analyzed in detail, focusing on two key aspects: the presence of data noise and the epoch-wise evolution of the internal structure. By leveraging this simple model architecture, we revealed the epoch-wise transitions of learning phases, internal signal separation between clean and noisy training data, and emergence of ``large activation'' observed in recent LLMs. These findings indicate that the deep double descent state occurring in the noisy data corresponds to the benign overfitting state, connecting concepts that have gained prominence in recent studies on machine learning and suggesting a novel, simple scenario for deep double descent. Results also indicated that sufficiently large and well-trained deep neural networks naturally achieve ``benign overfitting'' to data noise. This enables them to achieve strong adaptability for real-world tasks that inherently contain unavoidable noise.

\subsubsection*{Acknowledgments}

This work was supported by JSPS KAKENHI Grant Number JP25K07366 and by Grant Program of UNION TOOL Co. Scholarship Foundation.

\printbibliography

@article{belkin2019,
    title = {Reconciling modern machine-learning practice and the classical bias–variance trade-off},
    author = {Mikhail Belkin and Daniel Hsu and Siyuan Ma and Soumik Mandal},
    journal = {Proceedings of the National Academy of Sciences},
    volume = {116},
    number = {32},
    pages = {15849--15854},
    year = {2019},
}

@article{belkin2020,
    author = {Belkin, Mikhail and Hsu, Daniel and Xu, Ji},
    title = {Two Models of Double Descent for Weak Features},
    journal = {SIAM Journal on Mathematics of Data Science},
    volume = {2},
    number = {4},
    pages = {1167-1180},
    year = {2020},
    doi = {10.1137/20M1336072},
    URL = {https://doi.org/10.1137/20M1336072},
    eprint = {https://doi.org/10.1137/20M1336072}
}

@article{nakkiran2021a,
    title = {Deep double descent: where bigger models and more data hurt},
    author = {Nakkiran, Preetum and Kaplun, Gal and Bansal, Yamini and Yang, Tristan and Barak, Boaz and Sutskever, Ilya},
    journal = {Journal of Statistical Mechanics: Theory and Experiment},
    volume = {2021},
    number = {12},
    pages = {124003},
    year = {2021},
}

@article{shi2024,
    author = {Cheng Shi  and Liming Pan  and Hong Hu  and Ivan Dokmanić },
    title = {Homophily modulates double descent generalization in graph convolution networks},
    journal = {Proceedings of the National Academy of Sciences},
    volume = {121},
    number = {8},
    pages = {e2309504121},
    year = {2024},
    doi = {10.1073/pnas.2309504121},
    URL = {https://www.pnas.org/doi/abs/10.1073/pnas.2309504121},
    eprint = {https://www.pnas.org/doi/pdf/10.1073/pnas.2309504121}
}

@inproceedings{ascoli2020,
    author = {d\textquotesingle Ascoli, St\'{e}phane and Sagun, Levent and Biroli, Giulio},
    booktitle = {Advances in Neural Information Processing Systems},
    editor = {H. Larochelle and M. Ranzato and R. Hadsell and M.F. Balcan and H. Lin},
    pages = {3058--3069},
    publisher = {Curran Associates, Inc.},
    title = {Triple descent and the two kinds of overfitting: where \& why do they appear?},
    url = {https://proceedings.neurips.cc/paper_files/paper/2020/file/1fd09c5f59a8ff35d499c0ee25a1d47e-Paper.pdf},
    volume = {33},
    year = {2020}
}

@inproceedings{chen2021,
    author = {Chen, Lin and Min, Yifei and Belkin, Mikhail and Karbasi, Amin},
    booktitle = {Advances in Neural Information Processing Systems},
    editor = {M. Ranzato and A. Beygelzimer and Y. Dauphin and P.S. Liang and J. Wortman Vaughan},
    pages = {8898--8912},
    publisher = {Curran Associates, Inc.},
    title = {Multiple Descent: Design Your Own Generalization Curve},
    url = {https://proceedings.neurips.cc/paper_files/paper/2021/file/4ae67a7dd7e491f8fb6f9ea0cf25dfdb-Paper.pdf},
    volume = {34},
    year = {2021}
}

@inproceedings{bodin2021,
    author = {Bodin, Antoine and Macris, Nicolas},
    booktitle = {Advances in Neural Information Processing Systems},
    editor = {M. Ranzato and A. Beygelzimer and Y. Dauphin and P.S. Liang and J. Wortman Vaughan},
    pages = {21605--21617},
    publisher = {Curran Associates, Inc.},
    title = {Model, sample, and epoch-wise descents: exact solution of gradient flow in the random feature model},
    url = {https://proceedings.neurips.cc/paper_files/paper/2021/file/b4f8e5c5fb53f5ba81072451531d5460-Paper.pdf},
    volume = {34},
    year = {2021}
}

@misc{nakkiran2021b,
    title={Optimal Regularization Can Mitigate Double Descent}, 
    author={Preetum Nakkiran and Prayaag Venkat and Sham Kakade and Tengyu Ma},
    year={2021},
    eprint={2003.01897},
    archivePrefix={arXiv},
    primaryClass={cs.LG},
    url={https://arxiv.org/abs/2003.01897}, 
}

@article{geman1992,
    author = {Geman, Stuart and Bienenstock, Elie and Doursat, René},
    journal = {Neural Computation}, 
    title = {Neural Networks and the Bias/Variance Dilemma}, 
    year = {1992},
    volume = {4},
    number = {1},
    pages = {1-58},
    doi = {10.1162/neco.1992.4.1.1}
}

@inproceedings{yang2020,
    title = {Rethinking Bias-Variance Trade-off for Generalization of Neural Networks},
    author = {Yang, Zitong and Yu, Yaodong and You, Chong and Steinhardt, Jacob and Ma, Yi},
    booktitle = {Proceedings of the 37th International Conference on Machine Learning},
    pages = {10767--10777},
    year = {2020},
    editor = {III, Hal Daumé and Singh, Aarti},
    volume = {119},
    series = {Proceedings of Machine Learning Research},
    month = {7},
    publisher = {PMLR},
    url = {https://proceedings.mlr.press/v119/yang20j.html},
}

@inproceedings{adlam2020,
    author = {Adlam, Ben and Pennington, Jeffrey},
    booktitle = {Advances in Neural Information Processing Systems},
    editor = {H. Larochelle and M. Ranzato and R. Hadsell and M.F. Balcan and H. Lin},
    pages = {11022--11032},
    publisher = {Curran Associates, Inc.},
    title = {Understanding Double Descent Requires A Fine-Grained Bias-Variance Decomposition},
    url = {https://proceedings.neurips.cc/paper_files/paper/2020/file/7d420e2b2939762031eed0447a9be19f-Paper.pdf},
    volume = {33},
    year = {2020}
}

@inproceedings{chen2024,
    title = {On Bias-Variance Alignment in Deep Models},
    author = {Lin Chen and Michal Lukasik and Wittawat Jitkrittum and Chong You and Sanjiv Kumar},
    booktitle = {The Twelfth International Conference on Learning Representations},
    year = {2024},
    url = {https://openreview.net/forum?id=i2Phucne30}
}

@article{vallet1989,
    doi = {10.1209/0295-5075/9/4/003},
    url = {https://doi.org/10.1209/0295-5075/9/4/003},
    year = {1989},
    month = {6},
    volume = {9},
    number = {4},
    pages = {315},
    author = {F. Vallet and J.-G. Cailton and Ph Refregier},
    title = {Linear and Nonlinear Extension of the Pseudo-Inverse Solution for Learning Boolean Functions},
    journal = {Europhysics Letters},
}

@inproceedings{holzmüller2021,
    title = {On the Universality of the Double Descent Peak in Ridgeless Regression},
    author = {David Holzm{\"u}ller},
    booktitle = {International Conference on Learning Representations},
    year = {2021},
    url = {https://openreview.net/forum?id=0IO5VdnSAaH}
}

@article{mei2022,
    author = {Mei, Song and Montanari, Andrea},
    title = {The Generalization Error of Random Features Regression: Precise Asymptotics and the Double Descent Curve},
    journal = {Communications on Pure and Applied Mathematics},
    volume = {75},
    number = {4},
    pages = {667-766},
    doi = {https://doi.org/10.1002/cpa.22008},
    url = {https://onlinelibrary.wiley.com/doi/abs/10.1002/cpa.22008},
    eprint = {https://onlinelibrary.wiley.com/doi/pdf/10.1002/cpa.22008},
    year = {2022}
}

@article{hastie2022,
    author = {Trevor Hastie and Andrea Montanari and Saharon Rosset and Ryan J. Tibshirani},
    title = {{Surprises in high-dimensional ridgeless least squares interpolation}},
    volume = {50},
    journal = {The Annals of Statistics},
    number = {2},
    publisher = {Institute of Mathematical Statistics},
    pages = {949 -- 986},
    year = {2022},
    doi = {10.1214/21-AOS2133},
    URL = {https://doi.org/10.1214/21-AOS2133}
}

@article{bach2024,
    author = {Bach, Francis},
    title = {High-Dimensional Analysis of Double Descent for Linear Regression with Random Projections},
    journal = {SIAM Journal on Mathematics of Data Science},
    volume = {6},
    number = {1},
    pages = {26-50},
    year = {2024},
    doi = {10.1137/23M1558781},
    URL = {https://doi.org/10.1137/23M1558781},
    eprint = {https://doi.org/10.1137/23M1558781},
}

@article{deng2021,
    author = {Deng, Zeyu and Kammoun, Abla and Thrampoulidis, Christos},
    title = {A model of double descent for high-dimensional binary linear classification},
    journal = {Information and Inference: A Journal of the IMA},
    volume = {11},
    number = {2},
    pages = {435-495},
    year = {2021},
    month = {04},
    issn = {2049-8772},
    doi = {10.1093/imaiai/iaab002},
    url = {https://doi.org/10.1093/imaiai/iaab002},
    eprint = {https://academic.oup.com/imaiai/article-pdf/11/2/435/44020681/iaab002.pdf},
}

@misc{ullah2024,
    title={On the effect of noise on fitting linear regression models}, 
    author={Insha Ullah and A. H. Welsh},
    year={2024},
    eprint={2408.07914},
    archivePrefix={arXiv},
    primaryClass={math.ST},
    url={https://arxiv.org/abs/2408.07914}, 
}

@inproceedings{pezeshki2022,
    author = {Pezeshki, Mohammad and Mitra, Amartya and Bengio, Yoshua and Lajoie, Guillaume},
    title = {Multi-scale Feature Learning Dynamics: Insights for Double Descent},
    booktitle = {Proceedings of the 39th International Conference on Machine Learning},
    pages = {17669--17690},
    volume = {162},
    year = {2022},
    publisher = {PMLR}
}

@misc{stephenson2021,
    title={When and how epochwise double descent happens}, 
    author={Cory Stephenson and Tyler Lee},
    year={2021},
    eprint={2108.12006},
    archivePrefix={arXiv},
    primaryClass={cs.LG},
    url={https://arxiv.org/abs/2108.12006}, 
}

@inproceedings{heckel2021,
    title = {Early Stopping in Deep Networks: Double Descent and How to Eliminate it},
    author = {Reinhard Heckel and Fatih Furkan Yilmaz},
    booktitle = {International Conference on Learning Representations},
    year = {2021},
    url = {https://openreview.net/forum?id=tlV90jvZbw}
}

@inproceedings{gamba2023,
    author = {Matteo Gamba and Hossein Azizpour and Marten Bjorkman},
    title = {On the Lipschitz Constant of Deep Networks and Double Descent},
    booktitle = {34th British Machine Vision Conference 2023, {BMVC} 2023, Aberdeen, UK, November 20-24, 2023},
    publisher = {BMVA},
    year = {2023},
    url = {https://papers.bmvc2023.org/0871.pdf}
}

@misc{power2022,
    title={Grokking: Generalization Beyond Overfitting on Small Algorithmic Datasets}, 
    author={Alethea Power and Yuri Burda and Harri Edwards and Igor Babuschkin and Vedant Misra},
    year={2022},
    eprint={2201.02177},
    archivePrefix={arXiv},
    primaryClass={cs.LG},
    url={https://arxiv.org/abs/2201.02177}, 
}

@InProceedings{humayun2024,
    title = {Deep Networks Always Grok and Here is Why},
    author = {Humayun, Ahmed Imtiaz and Balestriero, Randall and Baraniuk, Richard},
    booktitle = {Proceedings of the 41st International Conference on Machine Learning},
    pages = {20722--20745},
    year = {2024},
    editor = {Salakhutdinov, Ruslan and Kolter, Zico and Heller, Katherine and Weller, Adrian and Oliver, Nuria and Scarlett, Jonathan and Berkenkamp, Felix},
    volume = {235},
    series = {Proceedings of Machine Learning Research},
    month = {7},
    publisher = {PMLR},
    url = {https://proceedings.mlr.press/v235/humayun24a.html},
}

@misc{li2025,
    title={Grokking in LLM Pretraining? Monitor Memorization-to-Generalization without Test}, 
    author={Ziyue Li and Chenrui Fan and Tianyi Zhou},
    year={2025},
    eprint={2506.21551},
    archivePrefix={arXiv},
    primaryClass={cs.LG},
    url={https://arxiv.org/abs/2506.21551}, 
}

@inproceedings{liu2023,
    title = {Omnigrok: Grokking Beyond Algorithmic Data},
    author = {Ziming Liu and Eric J Michaud and Max Tegmark},
    booktitle = {The Eleventh International Conference on Learning Representations },
    year = {2023},
    url = {https://openreview.net/forum?id=zDiHoIWa0q1}
}

@inproceedings{davies2022,
    title = {Unifying Grokking and Double Descent},
    author = {Xander Davies and Lauro Langosco and David Krueger},
    booktitle = {NeurIPS ML Safety Workshop},
    year = {2022},
    url = {https://openreview.net/forum?id=JqtHMZtqWm}
}

@misc{borkar2025,
    title={A dynamic view of some anomalous phenomena in SGD}, 
    author={Vivek Shripad Borkar},
    year={2025},
    eprint={2505.01751},
    archivePrefix={arXiv},
    primaryClass={math.OC},
    url={https://arxiv.org/abs/2505.01751}, 
}

@inproceedings{neyshabur2018,
    title = {The role of over-parametrization in generalization of neural networks},
    author = {Behnam Neyshabur and Zhiyuan Li and Srinadh Bhojanapalli and Yann LeCun and Nathan Srebro},
    booktitle = {International Conference on Learning Representations},
    year = {2019},
    url = {https://openreview.net/forum?id=BygfghAcYX},
}

@article{bartlett2020,
    author = {Peter L. Bartlett  and Philip M. Long  and Gábor Lugosi  and Alexander Tsigler},
    title = {Benign overfitting in linear regression},
    journal = {Proceedings of the National Academy of Sciences},
    volume = {117},
    number = {48},
    pages = {30063-30070},
    year = {2020},
    doi = {10.1073/pnas.1907378117},
    URL = {https://www.pnas.org/doi/abs/10.1073/pnas.1907378117},
    eprint = {https://www.pnas.org/doi/pdf/10.1073/pnas.1907378117},
}

@article{tsigler2023,
    author = {Alexander Tsigler and Peter L. Bartlett},
    title = {Benign overfitting in ridge regression},
    journal = {Journal of Machine Learning Research},
    year = {2023},
    volume = {24},
    number = {123},
    pages = {1--76},
    url = {http://jmlr.org/papers/v24/22-1398.html}
}

@inproceedings{xu2023,
    title = {Benign Overfitting and Grokking in {R}e{LU} Networks for {XOR} Cluster Data},
    author = {Zhiwei Xu and Yutong Wang and Spencer Frei and Gal Vardi and Wei Hu},
    booktitle = {NeurIPS 2023 Workshop on Mathematics of Modern Machine Learning},
    year = {2023},
    url = {https://openreview.net/forum?id=WvZV3JvmeR}
}

@inproceedings{cao2022,
    author = {Cao, Yuan and Chen, Zixiang and Belkin, Misha and Gu, Quanquan},
    booktitle = {Advances in Neural Information Processing Systems},
    editor = {S. Koyejo and S. Mohamed and A. Agarwal and D. Belgrave and K. Cho and A. Oh},
    pages = {25237--25250},
    publisher = {Curran Associates, Inc.},
    title = {Benign Overfitting in Two-layer Convolutional Neural Networks},
    url = {https://proceedings.neurips.cc/paper_files/paper/2022/file/a12c999be280372b157294e72a4bbc8b-Paper-Conference.pdf},
    volume = {35},
    year = {2022}
}

@InProceedings{frei2023,
    title = {Benign Overfitting in Linear Classifiers and Leaky {R}e{LU} Networks from {KKT} Conditions for Margin Maximization},
    author = {Frei, Spencer and Vardi, Gal and Bartlett, Peter and Srebro, Nathan},
    booktitle = {Proceedings of Thirty Sixth Conference on Learning Theory},
    pages = {3173--3228},
    year = {2023},
    editor = {Neu, Gergely and Rosasco, Lorenzo},
    volume = {195},
    series = {Proceedings of Machine Learning Research},
    month = {7},
    publisher = {PMLR},
    url = {https://proceedings.mlr.press/v195/frei23a.html},
}

@InProceedings{kou2023,
    title = {Benign Overfitting in Two-layer {R}e{LU} Convolutional Neural Networks},
    author = {Kou, Yiwen and Chen, Zixiang and Chen, Yuanzhou and Gu, Quanquan},
    booktitle = {Proceedings of the 40th International Conference on Machine Learning},
    pages = {17615--17659},
    year = {2023},
    editor = {Krause, Andreas and Brunskill, Emma and Cho, Kyunghyun and Engelhardt, Barbara and Sabato, Sivan and Scarlett, Jonathan},
    volume = {202},
    series = {Proceedings of Machine Learning Research},
    month = {7},
    publisher = {PMLR},
    url = {https://proceedings.mlr.press/v202/kou23a.html},
}

@misc{jones2025,
    title={Generalisation and benign over-fitting for linear regression onto random functional covariates}, 
    author={Andrew Jones and Nick Whiteley},
    year={2025},
    eprint={2508.13895},
    archivePrefix={arXiv},
    primaryClass={stat.ML},
    url={https://arxiv.org/abs/2508.13895}, 
}

@inproceedings{wilson2025,
    title = {Deep Learning is Not So Mysterious or Different},
    author = {Andrew Gordon Wilson},
    booktitle = {Forty-second International Conference on Machine Learning Position Paper Track},
    year = {2025},
    url = {https://openreview.net/forum?id=42Au7FoD8F}
}

@inproceedings{huang2024,
    title = {Unified View of Grokking, Double Descent and Emergent Abilities: A Comprehensive Study on Algorithm Task},
    author = {Yufei Huang and Shengding Hu and Xu Han and Zhiyuan Liu and Maosong Sun},
    booktitle = {First Conference on Language Modeling},
    year = {2024},
    url = {https://openreview.net/forum?id=cG1EbmWiSs}
}

@Techreport{krizhevsky2009,
    author = {Alex Krizhevsky and Vinod Nair and Geoffrey Hinton},
    title = {Learning Multiple Layers of Features from Tiny Images},
    institution = {University of Toronto},
    year = {2009},
    url = {https://www.cs.toronto.edu/~kriz/learning-features-2009-TR.pdf}
}

@inproceedings{kingma2015,
    author = {Diederik P. Kingma and Jimmy Ba},
    editor = {Yoshua Bengio and Yann LeCun},
    title = {Adam: {A} Method for Stochastic Optimization},
    booktitle = {3rd International Conference on Learning Representations, {ICLR} 2015, San Diego, CA, USA, May 7-9, 2015, Conference Track Proceedings},
    year = {2015},
    url = {http://arxiv.org/abs/1412.6980},
    bibsource = {dblp computer science bibliography, https://dblp.org}
}

@InProceedings{glorot2011,
    title = {Deep Sparse Rectifier Neural Networks},
    author = {Glorot, Xavier and Bordes, Antoine and Bengio, Yoshua},
    booktitle = {Proceedings of the Fourteenth International Conference on Artificial Intelligence and Statistics},
    pages = {315--323},
    year = {2011},
    editor = {Gordon, Geoffrey and Dunson, David and Dudík, Miroslav},
    volume = {15},
    series = {Proceedings of Machine Learning Research},
    address = {Fort Lauderdale, FL, USA},
    month = {4},
    publisher = {PMLR},
}

@InProceedings{bridle1990,
    author = {Bridle, John S.},
    title = {Probabilistic Interpretation of Feedforward Classification Network Outputs, with Relationships to Statistical Pattern Recognition},
    booktitle = {Neurocomputing},
    year = {1990},
    publisher = {Springer Berlin Heidelberg},
    address = {Berlin, Heidelberg},
    pages = {227--236},
}

@InProceedings{arpit2017,
    title = {A Closer Look at Memorization in Deep Networks},
    author = {Devansh Arpit and Stanis{\l}aw Jastrz{\k{e}}bski and Nicolas Ballas and David Krueger and Emmanuel Bengio and Maxinder S. Kanwal and Tegan Maharaj and Asja Fischer and Aaron Courville and Yoshua Bengio and Simon Lacoste-Julien},
    booktitle = {Proceedings of the 34th International Conference on Machine Learning},
    pages = {233--242},
    year = {2017},
    editor = {Precup, Doina and Teh, Yee Whye},
    volume = {70},
    series = {Proceedings of Machine Learning Research},
    month = {8},
    publisher = {PMLR},
    url = {https://proceedings.mlr.press/v70/arpit17a.html},
}

@inproceedings{liu2020,
    author = {Liu, Sheng and Niles-Weed, Jonathan and Razavian, Narges and Fernandez-Granda, Carlos},
    title = {Early-learning regularization prevents memorization of noisy labels},
    year = {2020},
    isbn = {9781713829546},
    publisher = {Curran Associates Inc.},
    address = {Red Hook, NY, USA},
    booktitle = {Proceedings of the 34th International Conference on Neural Information Processing Systems},
    articleno = {1707},
    numpages = {12},
    location = {Vancouver, BC, Canada},
    series = {NIPS '20}
}

@InProceedings{liu2022,
    title = {Robust Training under Label Noise by Over-parameterization},
    author = {Liu, Sheng and Zhu, Zhihui and Qu, Qing and You, Chong},
    booktitle = {Proceedings of the 39th International Conference on Machine Learning},
    pages = {14153--14172},
    year = {2022},
    editor = {Chaudhuri, Kamalika and Jegelka, Stefanie and Song, Le and Szepesvari, Csaba and Niu, Gang and Sabato, Sivan},
    volume = {162},
    series = {Proceedings of Machine Learning Research},
    month = {7},
    publisher = {PMLR},
    url = {https://proceedings.mlr.press/v162/liu22w.html},
}

@misc{priebe2022,
    title={Deep Learning is Provably Robust to Symmetric Label Noise}, 
    author={Carey E. Priebe and Ningyuan Huang and Soledad Villar and Cong Mu and Li Chen},
    year={2022},
    eprint={2210.15083},
    archivePrefix={arXiv},
    primaryClass={stat.ML},
    url={https://arxiv.org/abs/2210.15083}, 
}

@inproceedings{bondarenko2021,
    title = {Understanding and Overcoming the Challenges of Efficient Transformer Quantization},
    author = {Bondarenko, Yelysei  and Nagel, Markus  and Blankevoort, Tijmen},
    editor = {Moens, Marie-Francine  and Huang, Xuanjing  and Specia, Lucia  and Yih, Scott Wen-tau},
    booktitle = {Proceedings of the 2021 Conference on Empirical Methods in Natural Language Processing},
    month = {11},
    year = {2021},
    address = {Online and Punta Cana, Dominican Republic},
    publisher = {Association for Computational Linguistics},
    url = {https://aclanthology.org/2021.emnlp-main.627/},
    doi = {10.18653/v1/2021.emnlp-main.627},
    pages = {7947--7969},
}

@inproceedings{sun2024,
    title = {Massive Activations in Large Language Models},
    author = {Mingjie Sun and Xinlei Chen and J Zico Kolter and Zhuang Liu},
    booktitle = {ICLR 2024 Workshop on Mathematical and Empirical Understanding of Foundation Models},
    year = {2024},
    url = {https://openreview.net/forum?id=1ayU4fMqme}
}

@misc{yu2025,
    title={The Super Weight in Large Language Models}, 
    author={Mengxia Yu and De Wang and Qi Shan and Colorado J Reed and Alvin Wan},
    year={2025},
    eprint={2411.07191},
    archivePrefix={arXiv},
    primaryClass={cs.CL},
    url={https://arxiv.org/abs/2411.07191}, 
}

@inproceedings{zhang2017,
    title = {Understanding deep learning requires rethinking generalization},
    author = {Chiyuan Zhang and Samy Bengio and Moritz Hardt and Benjamin Recht and Oriol Vinyals},
    booktitle = {International Conference on Learning Representations},
    year = {2017},
    url = {https://openreview.net/forum?id=Sy8gdB9xx}
}

@inproceedings{gu2024,
    title = {Unraveling the Enigma of Double Descent: An In-depth Analysis through the Lens of Learned Feature Space},
    author = {Yufei Gu and Xiaoqing Zheng and Tomaso Aste},
    booktitle = {The Twelfth International Conference on Learning Representations},
    year = {2024},
    url = {https://openreview.net/forum?id=CEkIyshNbC}
}

\end{document}